\documentstyle[a4wide,named,11pt]{article}
\begin{document}
\begin{center}
{\Large\bf A FLEXIBLE SHALLOW APPROACH}
\end{center}
\begin{center}
{\Large\bf  TO TEXT GENERATION}
\end{center}
\vspace{0,5ex}
\begin{center}
Stephan Busemann and Helmut Horacek\linebreak
DFKI GmbH\linebreak
Stuhlsatzenhausweg 3, 66123 Saarbr\"ucken, Germany\linebreak
{\tt \{busemann, horacek\}@dfki.de}\footnote{This work has been supported by
a grant for the project TEMSIS from the European Union (Telematics
Applications Programme, Sector C9, contract no.\ 2945).}  
\end{center}
\vspace{0,5ex}
{\small
\begin{abstract}
In order to support the efficient development of NL
generation systems, two orthogonal methods are currently pursued with emphasis:
(1) reusable, general, and linguistically motivated surface realization
components, and (2) simple, task-oriented template-based techniques.
In this paper we argue that, from an application-oriented
perspective, the benefits of both are still limited. In order to improve this
situation, we suggest and evaluate shallow
generation methods associated with increased flexibility.
We advise a close connection between domain-motivated and
linguistic ontologies that supports the quick adaptation to new tasks
and domains, rather than the reuse of general resources.
Our method is especially designed for generating reports with
limited linguistic variations.
\end{abstract}
}
\section{Introduction}

In order to support the efficient development of NL
generation systems, two orthogonal methods are currently pursued with
emphasis: (1) reusable, general, and linguistically motivated surface
realization components, and (2) simple, task-oriented template-based
techniques. Surface realization components impose a layer of intermediate
representations that has become fairly standard, such as
the Sentence Plan Language (SPL) \cite{Kas:Whi:89}.
This layer allows for the use of existing software with well-defined 
interfaces, often reducing the development effort for surface realization
considerably.
Template-based techniques recently had some sort of revival through several
application-oriented projects such as {\sc idas} \cite{Rei:Mel:Lev:95},
that combine pre-defined surface expressions with freely generated
text in one or another way. 
However, the benefits of both surface realization components
and template-based techniques are still limited from an
application-oriented perspective. Surface realization components are
difficult to use because of the differences between 
domain-oriented and linguistically motivated ontologies (as in SPL), and
existing template-based techniques are too inflexible.

In this paper we suggest and evaluate flexible shallow methods for
report generation applications requiring limited linguistic resources
that are adaptable with little effort. We advise a close connection
between domain-motivated and linguistic 
ontologies, and we suggest a layer of intermediate representation that
is oriented towards the domain and the given task. This layer may
contain representations of different granularity, some highly implicit,
others very elaborate. We show how this is used by the processing
components in a beneficial way.  

The approach suggested does not only change the modularization
generally assumed for NLG systems drastically, it also renders the
system much more application-dependent. At first glance, however, such
an approach seems to abandon generality and reusability
completely, but, as we will demonstrate, this is not necessarily the
case. 

The rest of this paper is organized as follows: Section~\ref{related}
identifies deficits with current approaches to surface realization that
may occur for particular applications. In Section~\ref{work} we
propose alternative methods implemented into our sample application,
the generation of air-quality reports from current environmental
data. In Section~\ref{eval} we discuss the pros and cons of our
approach, and we summarize the conditions for successful use. 

\section{In-Depth and Shallow Generation}
\label{related}

\subsection{Shallow generation}
Recently, the distinction between in-depth and shallow approaches to language
processing has emerged from the need to build sensible
applications. In language understanding 
{\em deep} analysis
attempts to ``understand'' every part of the input, 
while {\em shallow}
analysis tries to identify only parts of interest for a particular
application.
Shallow analysis is
a key concept for information extraction from huge text bases and many
other real-world application types.

In language generation a corresponding distinction which we term {\em
in-depth\/} vs.\ {\em shallow genera\-tion}\footnote{We thus avoid
confusion with the common distinction 
between deep and surface generation.}  is becoming
prominent. While in-depth generation is inherently
knowledge-based and theoretically motivated,  
shallow generation  quite opportunistically models only the parts of
interest for the application in hand. Often such models will turn out
to be extremely shallow and simple, but in other cases much more detail
is required. Thus, developing techniques for varying modeling
granularity according to the requirements posed by the  application is
a prerequisite for more custom-tailored systems.  

According to Reiter and Mellish, shallow techniques (which they call
``intermediate'') are appropriate as long as corresponding in-depth
approaches are poorly understood, less efficient, or more costly to
develop \cite{Rei:Mel:93}. While our motivation for  shallow techniques
is in essence based on the cost factor, our assessment is even more
pronounced than Reiter's and Mellish's in that we claim that shallow
approaches combining different granularity in a flexible way
are better suited for small applications. We are convinced that shallow
generation systems will have a similar impact on the development of feasible
applications as shallow analyzers. 

\subsection{Potential shortcomings of approaches to surface
realization} 
Current approaches to surface realization are mostly in-depth, 
based on general, linguistically motivated, and widely reusable realization
components, such as Penman \cite{Penman:89}, KPML \cite{Bateman:97}, 
and SURGE \cite{Elh:Rob:96}.
These components are domain-independent and based on sound
linguistic principles. KPML and SURGE also exhibit a
broad coverage of English, while several other language models
are also available or under development. 
Despite their being reusable in general, the fact that the
modularization of grammatical
knowledge follows linguistic criteria rather than 
the needs of 
different types of applications may cause a number of
problems for an efficient development of concrete applications:

\begin{itemize}

\item The substantial differences
between domain- and linguistically motivated ontologies may
render the mapping between them difficult; for instance,
the use of case relations such as ``agent''
or ``objective'' requires compatible models of 
deep case semantics.

\item 
The need to encapsulate grammar
knowledge within the surface realizer
may require details in the intermediate representation
to be spelled out that are irrelevant
to the intended application,
even for rather small systems.

\item The fixed granularity of grammatical modeling requires a realizer
to cover many more languages, language fragments, or stylistic
variations than would be needed for one particular application,
which can lead to a considerable inefficiency of the realizer. 

\end{itemize}

In addition, there may be linguistic constructs needed for some
applications that are still outside the scope of the general tool. Their
inclusion may require the intermediate representation layer to be modified.

\subsection{Potential shortcomings of shallow generation methods}

A prominent example for an early shallow generation  system  is Ana
\cite{Kukich:83}, which reports about stock market developments.
While the kind of texts it produces can still be considered valuable 
today, Ana is implemented as a widely unstructured rule-based system, which
does not seem to be easily extendable and portable. Since then, 
various shallow methods including canned text parts and some template-based
techniques have been 
utilized, e.g.\ in CogentHelp \cite{Whi:Cal:97}, in the system described
in \cite{Caw:Bin:Jon:95},
and in {\sc idas} \cite{Rei:Mel:Lev:95}. They feature simplicity where the
intended application does not  
require fine-grained distinctions, such as the following techniques
used in {\sc idas}: 
\begin{itemize}
\item canned text with embedded KB references
(``Carefully slide [x] out along its guide''),
\item case frames with textual slot fillers, (``gently'' in  {\tt
(manner:~"gently")}). 
\end{itemize}
Although these techniques seem to be able to provide the necessary
distinctions for many practical applications in a much simpler way than
in-depth surface realization components can do,  a serious limitation
lies in their inflexibility. The first example above requires the
realization of [x] to agree in number with the canned part; as this is
not explicitly treated, the system seems to implicitly ``know'' that
only singular descriptions will be inserted.  
Moreover, canned texts as  case role fillers
may bear contextual influence, too, such as pronominals, or word
order phenomena. Thus, the flexibility of shallow generation
techniques should be increased
significantly.
\section{Shallow Generation in TEMSIS}
\label{work}
In order to tailor the design of a generation system towards an
application, we must account for different levels of granularity. We 
need a formalism capable of adapting to the expressivity of the
domain-oriented information. Parts of the texts to be generated may be
canned, some require templates, others require a more elaborate
grammatical model. 

In this section we first introduce an instance of the kind of
applications we have in mind.  %
We then proceed by discussing aspects of different granularity from the
point of view of the intermediate representation (IR) layer and the
components it interfaces. These include {\em text organization\/} and
{\em text realization}. The text organizer is also responsible for
content selection. It retrieves the relevant data from the TEMSIS
database. It combines fixed text blocks with the results of the
realizer in a language-neutral way. IR expressions are consumed by the
text realizer, which is a version of the production system TG/2
described in \cite{Busemann:96b}.  

\subsection{The TEMSIS application}

With~TEMSIS  a {\bf T}ransnational {\bf E}nvironmental {\bf 
M}anagement
{\bf S}upport and {\bf I}nformation {\bf S}ystem was created
as part of a transnational cooperation between the communities
in the French-German urban agglomeration, Moselle Est and Stadtverband
Saarbr\"ucken. Networked information kiosks are being installed in a number
of communities to provide public and expert environmental information. 
The timely availability of relevant environmental information 
will improve the planning and reactive capabilities
of the administration considerably. 
Current measurement data are made available on the TEMSIS web
server. The data include the pollutant, the measurement values, 
the location and the time the
measurements were taken, and a variety of thresholds.
Besides such data, the server provides metadata that allow for
descriptions of the measuring locations, of the pollutants measured and
of regulations or laws according
to which a comparison between measurements and thresholds can be
performed. This information can be accessed via the internet through
a hyperlink navigation interface ({\tt
http://www-temsis.dfki.uni-sb.de/}). 

The verbalization of NL air quality information in German and French
is an additional service reducing the need to look
up multiple heterogeneous data.  The generated texts can be
complemented with diagrams of time series. 
The material can be edited and further processed by the administrations
to fit additional needs. 

In order to request a report, a user specifies his demand by choosing
from a hierarchy of options presented to him within the hyperlink
navigation interface. He selects a report type by indicating whether he
is interested in average values, maximum values, or situations where
thresholds are exceeded.  Further choices include the language, the
country the environmental legislation of which should apply, the
measurement location, the pollutant, the period of time for which
measurements should be retrieved, and in some cases comparison
parameters. 
In addition, descriptions of pollutants and measurement stations can be
requested. They are stored as canned texts in the TEMSIS database. Not
all choices are needed in every case, and the TEMSIS navigator
restricts the combination of choices to the meaningful
ones. 

Let us assume that the user wants a French text comparing thresholds for sulfur
dioxide with measurements taken in the winter period of 1996/97 at V\"olklingen
City, and the applicable legislation should be from Germany. He also
wants a confirmation of some of his choices. 
The user receives the following text on his browser\footnote{A demo
version of the system is available at {\tt
http://www.dfki.de/service/nlg-demo/}.}  
(translated into English for the reader's convenience):

\begin{quotation}
\noindent {\sf You would like information about the concentration of sulfur
dioxide in the air during the winter season 1996/97. At the measurement
station of  V\"olklingen City, the early warning threshold 
for sulfur dioxide at an exposition of three hours (600
$\mu$g/m$^3$ according to the German decree ``Smogverordnung'') was not
exceeded.  In winter 1995/96, the early warning threshold was not
exceeded either.

\mbox{}
}
\end{quotation}

Reports are organized into one or several paragraphs.
Their length may range from a few lines to a page.

\begin{figure}[th]
{\small
\begin{verbatim}
        [(COOP THRESHOLD-EXCEEDING)
         (LANGUAGE FRENCH)                        
         (TIME [(PRED SEASON) (NAME [(SEASON WINTER) (YEAR 1996)])])
         (THRESHOLD-VALUE [(AMOUNT 600) (UNIT MKG-M3)])
         (POLLUTANT SULFUR-DIOXIDE)                          
         (SITE "V&o1lklingen-City")                
         (SOURCE [(LAW-NAME SMOGVERORDNUNG) (THRESHOLD-TYPE VORWARNSTUFE)]) 
         (DURATION [(HOUR 3)])                    
         (EXCEEDS [(STATUS NO) (TIMES 0)])]                   
\end{verbatim}
}

{\em En hiver 1996/97 \`a la station de mesure de V\"olklingen-City, 
le seuil d'avertissement pour le dioxide de soufre pour
une exposition de trois heures (600.0 $\mu$g/m$^3$ selon le decret
allemand ``Smogverordnung'') n'a pas \'et\'e d\'epass\'ee.}
\caption{A sample intermediate representation for a report
statement and its realization.}
\label{sample}
\end{figure}

\subsection{The intermediate representation}

The main purpose of the IR layer for the report generation system
consists in ensuring that all facets of the domain with their different
degrees of specificity can be verbally expressed, and in keeping the
realization task simple when no or little variety in language is
needed. While SPL and similar languages interfacing to
in-depth surface realization are either linguistic in nature or
largely constrain the surface form of an utterance, the IR specifies
domain information to be conveyed to the user and logical predicates
about it.  
Abstracting away from language-specific information in the IR like this
has the additional advantage that multi-lingual aspects can be kept
internal to the realizer. They depend on the {\tt LANGUAGE} feature in
an IR expression.  

The IR in Figure~\ref{sample} roughly corresponds to the key statement
of the sample report in the previous section (the second sentence),
which also appears at the end of each report as a summary. 
It constitutes a threshold comparison, as
stated by the value of the {\tt COOP}\footnote{The {\tt COOP} value
can correspond to the report type, as in the example, to confirmations
of user choices,  or to meta comments 
such as an introductory statement to a diagram, generated by a
dedicated component. } 
slot. There is only little indication 
as to how IR expressions should be expressed linguistically. 
Many semantic relations between
the elements of an IR expression are left implicit. 
For instance, the value of {\tt DURATION} relates to the
time of exposure according to the threshold's definition and not
to the period of time the user is interested in ({\tt
TIME}). Another example is the relation between {\tt EXCEEDS}
and {\tt THRESHOLD-VALUE}, which leads to the message that the early
warning threshold was not exceeded at all. Wordings are not
prescribed. For instance, our
sample IR does not contain a basis for the generation of ``exposure''
or ``measurement station''. 

IR expressions contain specifications at different degrees of granularity.
For coarse-grained specifications, it is up to the
text realizer to make missing or underspecified parts explicit on
the surface so that, in a sense, shallow text realization determines parts of
the contents. For more fine-grained  specifications, such as time expressions, 
text realization behaves like a general surface generator with 
a fully-detailed interface.
Ensuring an appropriate textual realization from IR expressions is 
left to the language template design within the realizer.

The syntax of IR expressions is defined by a standard Backus-Naur form. All
syntactically correct expressions have a compositional semantic 
interpretation and can be realized as a surface text provided corresponding
realization rules are defined.
Sharing the IR definitions between the text organization and the 
realization component thus avoids problems of realizability described
in \cite{Meteer:92}.

\subsection{Text organization}
\label{textorg}

The goal of text organization in our context
is to retrieve and express, in terms suitable for the definition of the
IR, (1) report specifications provided by the user, (2)
the relevant domain data accessed from the database
according to these specifications, 
including e.g.\ explicit comparisons between measurements and threshold
values, and (3) implicitly associated meta-information from the
database, such as the 
duration of exposure, the decree and the value of the threshold.
This task is accomplished by a staged process
that is application-oriented rather than  based on linguistically
motivated principles.

The process starts with building some sort of a representation 
sketch,
by instantiating a report skeleton that consists of a sequence
of assertion statement specifications.
Assertion statements consist of a top level predicate that represents
the assertion's type (e.g.\ threshold-exceeding) and encapsulates the
entire meaning of the associated assertion, except to attached
specifications and domain data, to  
make local parameters and data dependencies explicit.

In order to transform this initial representation to meet the 
application-oriented requirements of the IR,
it is necessary to recast the information, which comprises
{\em augmenting}, {\em restructuring}, and {\em aggregating} its components.

{\em Augmenting} statement specifications means making information
implicitly contained or available elsewhere explicitly at the place
it is needed. This concerns reestablishing report-wide information, as
well as making locally entailed information accessible. An example for
the former is the number of diagrams copied into the introductory
statement to these diagrams. This treatment is much simpler than using
a reference generation algorithm, but it relies on knowing the number
of diagrams in advance. An example for the latter is the unit in which
the value of a measurement is expressed. 

{\em Restructuring} information imposes some manipulations on the
specifications obtained so far to rearrange the pieces of
information contained so that they meet the definition of the IR. The
associated operations include reifying an attribute as a structured
value and raising an embedded partial description. 
These operations are realized by mapping schemata similar to those
elaborated for linguistically motivated lexicalization
\cite{Horacek:96}. However, some of our schemata are purely
application-oriented and tailored to the domain, which manifests itself
in the larger 
size of the structures covered.

{\em Aggregation}, the last part of information recasting, 
comprises removing partial descriptions or adding
simple structures.
These operations are driven by a small set of declaratively
represented rules that access a discourse memory.
Most of the rules aim at avoiding repetitions of optional constituents
(e.g., temporal and locative information) over adjacent statements. 
For example,  the {\tt TIME} specification is elided in the second
sentence of our sample text, since the time specification in the first
sentence still applies. An example for adding a simple structure to an
IR expression is the insertion of a marker indicating a strong
correspondence between adjacent assertions, which gives rise to
inserting ``either'' in the sample text. Altogether, the underlying
rules are formulated to meet application particularities, such as
impacts of certain combinations of a value, a status, and a threshold
comparison outcome, rather than to capture linguistic principles.

\subsection{Text realization with TG/2}
\label{tg2}
 
TG/2 is a flexible and reusable application-oriented text realization
system that can be smoothly combined with deep generation processes. It
integrates canned text, templates, and context-free rules into a single
production-rule formalism and is thus extremely well suited for coping
with IR subexpressions of different granularity. 

TG/2 is based on production system techniques \cite{Dav:Kin:77} that
preserve the modularity of processing and linguistic knowledge.
Productions are applied through the familiar three-step 
processing
cycle: (i) identify the applicable rules, (ii) select a rule on the
basis of some conflict resolution mechanism, and (iii) apply that rule.
Productions are used to encode grammar rules in the
language TGL \cite{Busemann:96b}. A rule is applicable if its
preconditions are met. The TGL rule in Figure~\ref{tgl} is applicable to
input material as shown in Figure~\ref{sample}, because the {\tt
COOP} slot
matches, and there is information about the {\tt THRESHOLD-VALUE} available
(otherwise a different sentence pattern, and hence a different rule,
would be required).  

\begin{figure*}
{\small
\begin{verbatim}
(defproduction  threshold-exceeding "WU01"
  (:PRECOND (:CAT DECL
             :TEST ((coop-eq 'threshold-exceeding) (threshold-value-p)))
   :ACTIONS (:TEMPLATE  (:OPTRULE PPtime (get-param 'time))
                        (:OPTRULE SITEV (get-param 'site))
                        (:RULE THTYPE (self))
                        (:OPTRULE POLL (get-param 'pollutant))
                        (:OPTRULE DUR (get-param 'duration))
                        "(" (:RULE VAL (get-param 'threshold-value))
                            (:OPTRULE LAW (get-param 'law-name)) ") "
                        (:RULE EXCEEDS (get-param 'exceeds)) "."
             :CONSTRAINTS (:GENDER (THTYPE EXCEEDS) :EQ))))
\end{verbatim}
}
\caption{A TGL rule defining a sentence template for threshold
exceeding statements. }
\label{tgl}
\end{figure*}

TGL rules contain categories as in a context-free grammar,
which are used for rule selection (see below). The rule's
actions are carried out in a top-down, depth-first and left-to-right
manner. They include the activation of other rules ({\tt
:RULE, :OPTRULE}), the execution of a function, or the
return of an ASCII string as a (partial) result.
When selecting other rules by virtue of a category, the
relevant portion of the input structure for which a candidate rule
must pass its associated tests must be identified. The function {\tt
get-param} in Figure~\ref{tgl} yields the substructure of the current
input depicted by the argument. The first action
selects all rules with category {\tt PPtime}; the 
relevant substructure is the {\tt TIME} slot of an IR.

TGL rules are defined according to the IR syntax definitions. This
includes optional IR elements, many of which can simply be omitted
without disturbing fluency. In these cases,  optional rules 
({\tt OPTRULE}) are defined in TGL. Optional actions
are ignored if the input structure does not contain relevant
information. In certain cases, the omission of an IR element would
suggest a different sentence structure, which is accounted for by
defining alternative TGL rules with appropriate tests for the presence of
some IR element.
Agreement relations are encoded into TGL by virtue of a \mbox{PATR}
style feature percolation mechanism \cite{Shi:Usz:Per:83}. The rules
can be annotated by equations that either assert equality of a
feature's value at two or more constituents, or introduce a feature
value at a constituent. The constraint in Figure~\ref{tgl} requires the
categories {\tt THTYPE} and {\tt EXCEEDS} to agree in gender, thus
implementing a subject-participle agreement relation in French. This
general mechanism provides a considerable amount of flexibility and
goes beyond simple template filling techniques. 

A TGL rule is successfully applied if all actions are carried out.  The
rule returns the concatenation of the substrings produced by the
``template'' actions. If an action fails, backtracking can be invoked
flexibly and efficiently using memoization techniques
(see \cite{Busemann:96b}). 

\section{Costs and Benefits}
\label{eval}

As Reiter and Mellish note, the use of shallow techniques needs to be
justified through a cost-benefit analysis \cite{Rei:Mel:93}.  
We specify the range of possible applications our approach is useful
for, exemplified by the report generator developed for the TEMSIS
project.  

This application took an effort of about eight person months, part of
which were spent implementing interfaces to the TEMSIS server and to
the database, and for making ourselves acquainted with details of the
domain. 
The remaining time was spent on (1) the elicitation of user
requirements and the definition of a small text corpus, (2) the design
of IR according to the domain distinctions required for the corpus
texts, and (3) text organization, adaptation of TG/2 and grammar
development.  

The grammars comprise 105 rules for the German and 122 for the French
version. There are about twenty test predicates and IR access
functions, most of which are needed for both languages. The French
version was designed on the basis of the German one and took little
more than a week to implement. The system covers a total of 384
different report structures that differ in at least one linguistic
aspect.  
\subsection{Benefits}

Altogether, the development effort was very low. We believe that
reusing an in-depth surface generator for this task would not have
scored better. Our method has a number of advantages: 

(1) {\em Partial reusability}. Despite its domain-dependence, parts of
the system are reusable. The TG/2 interpreter has been adopted without
modifications. Moreover, a sub-grammar for time expressions in the domain of
appointment scheduling was reused with only minor extensions. 

(2) {\em Modeling flexibility}. Realization techniques of different
granularity (canned text, templates, context-free grammars) allow the
grammar writer to model general, linguistic knowledge as well as more
specific task and domain-oriented wordings.  

(3) {\em Processing speed}. Shallow processing is fast. In our system,
the average generation time of less than a second can almost be
neglected (the overall run-time is longer due to database access). 

(4) {\em Multi-lingual extensions}. Additional languages can be
included with little effort because the IR is neutral towards particular
languages. 

(5) {\em Variations in wording}.
Alternative formulations are easily integrated by defining conflicting
rules in TGL. These are ordered according to a set of criteria that
cause the system to prefer certain formulations to others
(cf.\ \cite{Busemann:96b}). Grammar rules leading to preferred
formulations are selected first from a conflict set of concurring
rules. The preference mechanisms will be used in a future version to
tailor the texts for administrative and public uses. 
\subsection{Costs}

As argued  above, the orientation towards the application task and
domain yields some important benefits. On the other hand, there are
limitations in reusability and flexibility:

(1) IR cannot be reused for other applications. The consequences for
the modules interfaced by IR, the text organizer and the text realizer,
are a loss in generality. Since both modules keep a generic
interpreter apart from partly domain-specific knowledge, the effort
of  transporting the components to new applications is, however,
restricted to modifying the knowledge sources.  

(2) By associating canned text with domain acts, TG/2 behaves in a
domain and task specific way. This keeps the flexibility in the
wording, which can only partly be influenced by the text organizer,
inherently lower than with in-depth approaches. 

\subsection{When does it pay off?}

We take it for granted that the TEMSIS generation application stands
for a class of comparable tasks that can be characterized as
follows. The generated texts are information-conveying reports in a
technical domain. The sublanguage allows for a rather straight-forward
mapping onto IR expressions, and IR expressions can be realized
in a context-independent way.  
For these kinds of applications, our methods provide sufficient
flexibility by omitting unnecessary or known information from both the
schemes and its IR expressions, and by including particles to increase
coherency. The reports could be generated in multiple languages. We
recommend the opportunistic use of shallow techniques for this type of
application.  

Our approach is not suitable for tasks involving deliberate sentence
planning, the careful choice of lexemes, or a sophisticated
distribution of information onto linguistic units. 
Such tasks would not be compatible with the loose coupling of our
components via IR. In addition, they would require complex tests to be
formulated in TGL rules, rendering the grammar rather obscure. Finally,
if the intended coverage of content is to be kept extensible or is not
known precisely enough at an early phase of development, the eventual
redesign of the intermediate structure and associated mapping rules for
text organization may severely limit the usefulness of our approach.
\section{Conclusion}

We have suggested shallow approaches to NL generation that are suited
for small applications requiring limited linguistic resources. While
these approaches ignore many theoretical insights gained through years
of NLG research and instead revive old techniques once criticized for
their lack of flexibility, they nevertheless allow for the quick
development of running systems. By integrating techniques of different
granularity into one formalism, we have shown that lack of flexibility
is not an inherent property of shallow approaches. Within the air
quality report generation in TEMSIS, a non-trivial application was
described. We also gave a qualitative evaluation of the domain
characteristics to be met for our approach to work
successfully. Further experience will show whether shallow techniques
transpose to more complex tasks. 

We consider it a scientific challenge
to combine shallow and in-depth approaches to analysis and generation
in such a way that more theoretically motivated research finds its way
into real applications.  

{\small
\bibliography{update,cl-general-cross,cl-pub-cross,cl-general,cl-pub}
}
\end{document}